\documentclass[letterpaper]{article} 
\usepackage{aaai24}  
\usepackage{times}  
\usepackage{helvet}  
\usepackage{courier}  
\usepackage[hyphens]{url}  
\usepackage{graphicx} 
\usepackage{natbib}  
\usepackage{caption} 
\usepackage{algorithm}
\usepackage{algorithmic}

\usepackage{url}
\usepackage{amsmath}
\usepackage{amssymb}
\usepackage{mathtools}
\usepackage{amsthm}
\usepackage{booktabs}

\usepackage{newfloat}
\usepackage{listings}
\DeclareCaptionStyle{ruled}{labelfont=normalfont,labelsep=colon,strut=off} 
\floatstyle{ruled}
\newfloat{listing}{tb}{lst}{}
\floatname{listing}{Listing}
\title{Imagine, Initialize, and Explore: An Effective Exploration Method in Multi-Agent Reinforcement Learning}
\author{
    Zeyang Liu, Lipeng Wan, Xinrui Yang, Zhuoran Chen, Xingyu Chen, Xuguang Lan\footnote{Corresponding author.}
}
\affiliations{
    National Key Laboratory of Human-Machine Hybrid Augmented Intelligence\\
    National Engineering Research Center for Visual Information and Application\\
    Institute of Artificial Intelligence and Robotics, Xi'an Jiaotong University, Xi'an, China, 710049\\
    \{zeyang.liu, wanlipeng, xinrui.yang, zhuoran.chen\}@stu.xjtu.edu.cn, xingyuchen1990@gmail.com, xglan@mail.xjtu.edu.cn
}

\begin{document}

\maketitle

\begin{abstract}
    Effective exploration is crucial to discovering optimal strategies for multi-agent reinforcement learning (MARL) in complex coordination tasks. Existing methods mainly utilize intrinsic rewards to enable committed exploration or use role-based learning for decomposing joint action spaces instead of directly conducting a collective search in the entire action-observation space. However, they often face challenges obtaining specific joint action sequences to reach successful states in long-horizon tasks. To address this limitation, we propose Imagine, Initialize, and Explore (IIE), a novel method that offers a promising solution for efficient multi-agent exploration in complex scenarios. IIE employs a transformer model to imagine how the agents reach a critical state that can influence each other's transition functions. Then, we initialize the environment at this state using a simulator before the exploration phase. We formulate the imagination as a sequence modeling problem, where the states, observations, prompts, actions, and rewards are predicted autoregressively. The prompt consists of timestep-to-go, return-to-go, influence value, and one-shot demonstration, specifying the desired state and trajectory as well as guiding the action generation. By initializing agents at the critical states, IIE significantly increases the likelihood of discovering potentially important under-explored regions. Despite its simplicity, empirical results demonstrate that our method outperforms multi-agent exploration baselines on the StarCraft Multi-Agent Challenge (SMAC) and SMACv2 environments. Particularly, IIE shows improved performance in the sparse-reward SMAC tasks and produces more effective curricula over the initialized states than other generative methods, such as CVAE-GAN and diffusion models.
\end{abstract}

\section{Introduction}

Recent progress in cooperative multi-agent reinforcement learning (MARL) has shown attractive prospects for real-world applications, such as autonomous driving~\citep{zhou2021smarts} and active voltage control on power distribution networks~\citep{wang2021multi}. To utilize global information during training and maintain scalability in execution, centralized training with decentralized execution (CTDE) has become a widely adopted paradigm in MARL. Under this paradigm, value decomposition methods~\citep{rashid2018qmix,son2019qtran,rashid2020weighted} factorize the joint $Q$-value as a function of individual utility functions, ensuring consistency between the centralized policy and the individual policies. Consequently, they have achieved state-of-the-art performance in challenging tasks, such as StarCraft unit micromanagement~\cite{samvelyan2019starcraft}.

\begin{figure*}[t]
    \centering
    \includegraphics[width=0.81\linewidth]{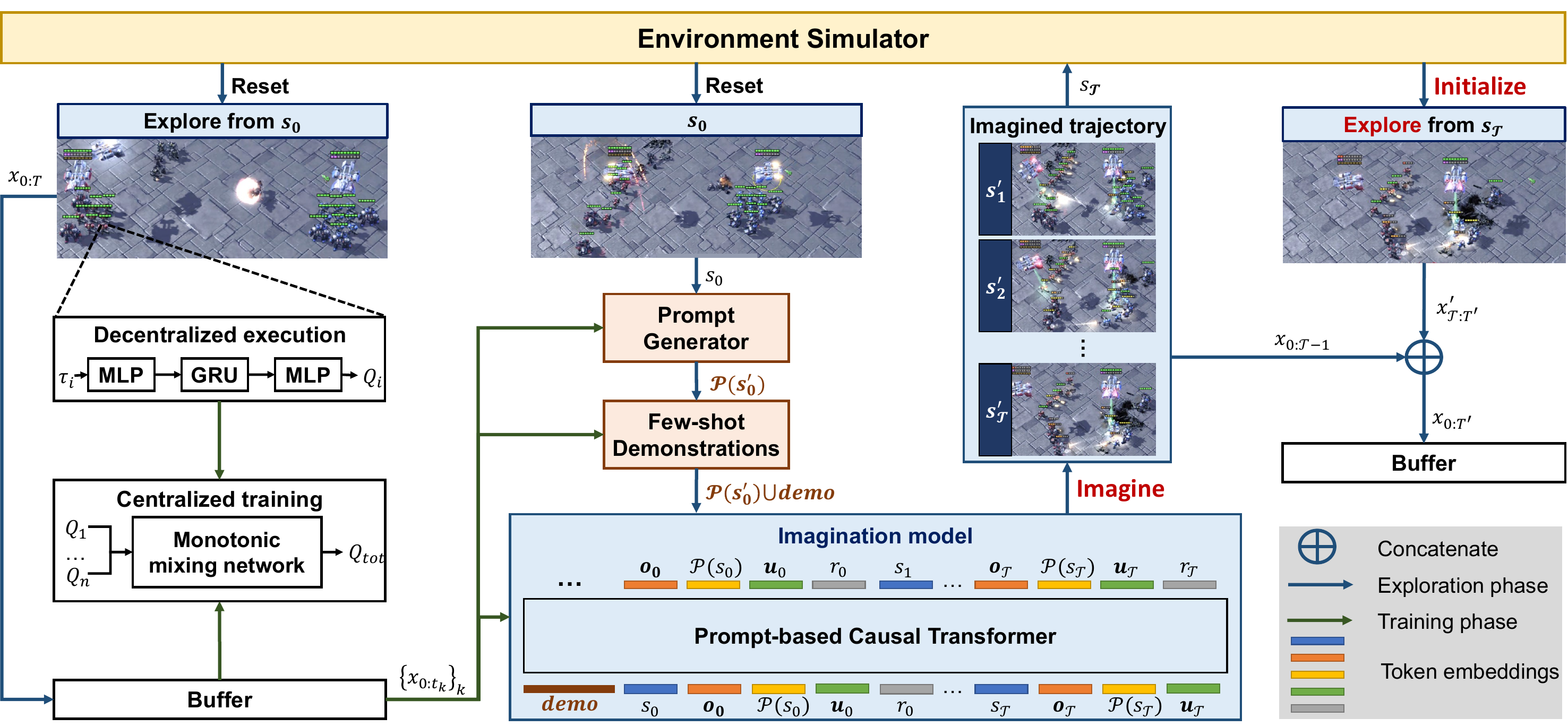}
    \caption{An overview of Imagine, Initialize, and Explore. In the pretraining phase, individual agents collect data from the initial state $s_0$ provided by the environment simulator. The interaction sequence is divided into several trajectory segments using influence values, which serve as the training dataset for the imagination model and few-shot demonstrations. Given $s_0$, the prompt generator is trained to produce critical states, and the imagination model learns to predict how to reach such critical states from $s_0$. After pretraining, the imagination model generates a trajectory from the initial state $s_0$ to a critical state $s_\mathcal{T}$ conditioned on $\mathcal{P}(s_0)$ sampled from the prompt generator, and the most related trajectory from the few-shot demonstration dataset. The agents are initialized at $s_\mathcal{T}$ by the environment simulator and then interact with the environment using the $\epsilon$-greedy strategy. We concatenate the imagined and the explored trajectory to train the joint policy in the centralized training phase.}
    \label{overview}
\end{figure*}

Despite their success, the simple $\epsilon$-greedy exploration strategy used in these methods has been found ineffective in solving coordination tasks with complex state and reward transitions~\citep{mahajan2019maven,Wang2020Influence,zheng2021episodic}. To address this limitation, MAVEN~\citep{mahajan2019maven} adopts a latent space for hierarchical control, allowing agents to condition their behavior on shared latent variables and enabling committed exploration. EITI and EDTI~\citep{Wang2020Influence} quantify and characterize the influence of one agent’s behavior on others using mutual information and the difference of expected returns, respectively. By optimizing EITI or EDTI objectives as intrinsic rewards, agents are encouraged to coordinate their exploration and learn policies that optimize team performance. EMC~\citep{zheng2021episodic}, on the other hand, uses prediction errors of individual $Q$-values to capture the novelty of states and the influence from other agents for coordinated exploration. Unfortunately, these methods suffer from the exponential growth of the action-observation space with the number of agents and become inefficient in long-horizon tasks.

It is possible to decompose the complex task rather than directly conducting collective searches across the entire action-observation space. To this end, RODE~\cite{wang2020rode} decomposes joint action spaces into restricted role action spaces by clustering actions based on their effects on the environment and other agents. Low-level role-based policies explore and learn within these restricted action-observation spaces, reducing execution and training complexity. However, RODE still struggles with long-term tasks as it is still unlikely to obtain a long sequence of specific actions to achieve successful states. The Go-Explore family of algorithms~\citep{ecoffet2019go,guo2020memory,ecoffet2021first} decomposes the exploration into two phases: returning to the previous state and then starting to explore. These methods store the high-scoring trajectories in an archive and return to sampled states from this archive by running a goal-conditioned policy. However, in the multi-agent field, agents should be encouraged to teleport to the states where interactions happen and may lead to critical under-explored regions. These approaches only consider the sparse and deceptive rewards in single-agent settings, making them impractical for MARL with complex reward and transition dependencies among cooperative agents.

Teleporting agents to interaction states that influence each other's transition function can significantly increase the likelihood of discovering potentially important yet rarely visited states and reduce the exploration space. However, there often exist multiple feasible but inefficient trajectories to reach such interaction states due to the abundance of agents' tactics and the compositional nature of their functionalities. In light of this, we propose a novel MARL exploration method named Imagine, Initialize, and Explore (IIE). It leverages the GPT architecture~\citep{radford2018improving} to imagine trajectories from the initial state to interaction states, acting as a powerful ``memorization engine'' that can generate diverse agent behaviors. We use target timestep-to-go, return-to-go, influence value, and a one-shot demonstration as prompts to specify the ``path'' of the imagined trajectory. The influence value is an advantage function that compares an agent's current action $Q$-value to a counterfactual baseline that marginalizes this agent. Specifically, we obtain several trajectory segments by dividing the exploration episode, where each segment starts at the initial state from the environment and ends at an interaction state with the highest influence value. Then, the GPT model learns to predict states, observations, prompts, actions, and rewards in an autoregressive manner on these segments. 

Fig.~\ref{overview} presents an overview of IIE architecture. Before the exploration phase, the agents are given a new initial state and few-shot demonstrations to construct the prompt and generate the imagined trajectory. The agents operating in partially observable environments can benefit from this imagination by utilizing it to initialize the recurrent neural network state. Then, the agents are initialized to the last state of the imagined trajectory by the simulator and interact with the environment to explore. IIE gradually provides high-influence starting points that can be viewed as auto-curricula that help agents collect crucial under-explored regions. To make the best use of the imagination, we also stitch it with the interaction sequence from exploration for policy training.

The main contributions of this paper are threefold: First, it introduces Imagine, Initialize, and Explore, which leverages the GPT architecture to imagine how the agents reach critical states before exploration. This method bridges sequence modeling and transformers with MARL instead of using GPT as a replacement for reinforcement learning algorithms~\citep{chen2021decision}. Second, empirical results demonstrate significant performance improvements of IIE on partially observable MARL benchmarks, including StarCraft Multi-Agent Challenge (SMAC) with dense and sparse reward settings as well as SMACv2. And third, guided by the target timestep-to-go, return-to-go, influence value, and a one-shot demonstration, the imagination model can produce more effective curricula and outperforms behavior cloning, CVAE-GAN, and classifier-guided diffusion.

\section{Background}
\paragraph{Decentralized Partially Observable Markov Decision Process.} A fully cooperative multi-agent task in the partially observable setting can be formulated as a Decentralized Partially Observable Markov Decision Process (Dec-POMDP)~\cite{oliehoek2016concise}, consisting of a tuple $ G=\langle A,S,\Omega,O,U,P,r,\gamma\rangle $, where $ a\in A\equiv\left\{1,\ldots,n\right\} $ is a set of agents, $S$ is a set of states, and $\mathrm{\Omega}$ is a set of joint observations. At each time step, each agent obtains its observation $ o\in\mathrm{\Omega} $ based on the observation function $ O\left(s,a\right):S\times A\rightarrow\mathrm{\Omega} $, and an action-observation history $ \tau_a \in T \equiv (\Omega\times U)^\ast $. Each agent $a$ chooses an action $u_a\in U$ by a stochastic policy $\pi_a\left(u_a\middle|\tau_a\right):T\times U\rightarrow\left[0,1\right]$, which forms a joint action $ \mathbf{u}\in\mathbf{U} $. It results in a joint reward $r(s,\mathbf{u})$ and a transit to the next state $s'\sim P(\cdot|s,\textbf{u})$. The formal objective function is to find the joint policy $\boldsymbol{\pi}$ that maximizes a joint action-value function $Q^{\boldsymbol{\pi}}(s_t, \textbf{u}_t)=r(s_t,\textbf{u}_t)+\gamma \mathbb{E}_{s'}\left[V^{\boldsymbol{\pi}}(s')\right]$, where $V^{\boldsymbol{\pi}}(s)=\mathbb{E}\left[\sum_{t=0}^\infty \gamma^t r_t|s_0=s,\boldsymbol{\pi}\right]$, and $\gamma\in [0,1)$ is a discounted factor.

\paragraph{Centralized Training with Decentralized Execution.} In this paradigm, agents' policies are trained with access to global information in a centralized way and executed only based on local histories in a decentralized way~\citep{kraemer2016multi}. One of the most significant challenges is to guarantee the consistency between the individual policies and the centralized policy, which is also known as Individual-Global Max~\citep{son2019qtran}:
\begin{equation}
    \mathop{\arg\max}_{\mathbf{u}} Q_{j}(s_t,\mathbf{u})=
    \begin{Bmatrix}
        \mathop{\arg\max}_{u^1} Q^1(s_t,u^1)\\
        ...\\
        \mathop{\arg\max}_{u^n} Q^n(s_t,u^n)
    \end{Bmatrix}   
\end{equation}

To address this problem, QMIX~\citep{rashid2018qmix} applies a state-dependent monotonic mixing network $f_s$ to combine per-agent $Q$-value functions with the joint $Q$-value function $Q_j(s,\mathbf{u})$. The restricted space of all $Q_{j}(s,\mathbf{u})$ that QMIX can be represented as:
\begin{equation}
    \mathcal{Q}^{m}:=\{Q_{j}| Q_{j}=f_s(Q^1(s,u^1),...,Q^n(s,u^n)) \},
\end{equation}
where $Q^a(s,u^a)\in \mathbb{R}$, $\frac{\partial f_s}{\partial Q^a}\geqslant 0 , \forall a\in A$.

\section{Method}
This section presents a novel multi-agent exploration method, Imagine, Initialize, and Explore, consisting of three key components: (1) the imagination model, which utilizes a transformer model to generate the trajectory representing how agents reach a target state autoregressively, (2) the prompt generator, which specifies the target state and trajectory for imagination, and (3) the environment simulator, which can teleport the agents to a state instantly. 

\subsection{Imagination Model} 
It has been found that agents operating in a partially observable environment can benefit from the action-observation history~\citep{hausknecht2015deep,karkus2017qmdp,rashid2018qmix}, e.g., a model that has a recall capability such as gated recurrent unit (GRU). Therefore, it is necessary to imagine the trajectory from the initial state to the selected interaction state. We formulate the problem of trajectory generation as a sequence modeling task, where the sequences have the following form:
\begin{equation}
    x=\{...,s_t, o^1_t, ..., o^n_t, \mathcal{P}(s_t), u^1_t, ..., u^n_t, r_t, s^{t+1},...\},
\end{equation}
where $t$ represents the timestep, $n$ is the number of agents, and $\mathcal{P}(s_t)$ is the prompt for action generation at the state $s_t$, with the definition provided in the next subsection.

We obtain the token embeddings for states, observations, prompts, actions, and rewards through a linear layer followed by layer normalization. Moreover, an embedding for each timestep is learned and added to each token. The transformer model processes the tokens and performs autoregressive modeling to predict future tokens.

The imagination model is trained to focus on interaction states where the agents can influence each other's transition function by prioritizing trajectories with a high-influence last state in the sampled batch. Specifically, we split interaction sequences from the sampled batch into $K$ segments, starting from the initial state by the simulator and ending at the states with top-$K$ high-influence levels. To ensure generalization across the entire explored regions, we also enforce all state-action pairs of the sampled batch to be assigned a minimum probability of $\frac{\lambda}{N}$, where $\lambda$ is a hyperparameter, and $N$ is the number of state-action pairs in the batch. The influence $\mathcal{I}$ is defined as an advantage function that compares the $Q$-value for the current action $u^a$ to a counterfactual baseline, marginalizing out $u^{a*}$ at a given state $s$:
\begin{equation}
    \mathcal{I}(s)=\max_{a\in A}\left\{Q_{j}(s,\boldsymbol{\tau},\textbf{u}) - \mathbb{E}_{u^{a*}} Q_{j}\left(s,\boldsymbol{\tau},(u^{a*},u^{-a})\right)\right\},
\end{equation}
where $Q_{j}(s,\boldsymbol{\tau},\textbf{u})=f_s(Q^1(\tau^1,u^1;\theta), ..., Q^n(\tau^n,u^n;\theta);\phi)$ represents the joint $Q$-value function, $-a$ denotes all agents $A$ except agent $a$, $f_s$ denotes the monotonic mixing network whose non-negative weights are generated by hyper-networks that take the state as input. 

The imagination model parameterized by $\psi$ is trained by:
\begin{equation}
\begin{aligned}
    L_m &=\sum_{t=1}^{T} \bigg[\log q^{\psi}(s_t|x_{<s_t}) + \log q^{\psi}(r_t|x_{<r_t})\\    
    & +\sum_{a=1}^{n}\left(\log q^\psi(u^a_t|x_{<u^a_t}) + \log q^\psi(o^a_t|s_t)\right)\bigg],
\end{aligned}
\end{equation}
where $T$ is the episode length. Since the observation is only related to the current state and the vision range of the agents, we filter out the historical memories in $x<o_{t^a}$ and use $s_t$ as the input of the observation model $q^\psi(o^a_t|s_t)$.

\subsection{Prompt For Imagination}
The primary objective of our imagination model is to identify critical states and imagine how to reach them from the initial point. However, multiple feasible trajectories exist to reach these states due to the diverse tactics and compositional nature of agents' functionalities. Despite their feasibility, many of these trajectories are highly inefficient for imagination. For example, agents may wander around their initial points before engaging the enemy, decreasing success rates within a limited episode length. Therefore, a prompt, serving as a condition for action generation, is necessary to specify the target trajectory.

Given the starting state $s_i$, we propose a prompt generator $\mathcal{P}^\xi(s_i)$ to predict the sequence $\{s_i,\mathcal{I}_i,\mathcal{T}_i,\mathcal{R}_i\}$, where $\mathcal{I}_i$ is the influence value, $\mathcal{T}_i$ is the timestep-to-go representing how quickly we can achieve the interaction state, and $\mathcal{R}_i=\sum_{t=i}^{i+\mathcal{T}}r_t$ is the return-to-go. 

As the training datasets for the imagination model contain a mixture of trajectory segments, directly sampling from the prompt generator is unlikely to produce the critical state consistently. Instead, we aim to control the prompt generator to produce frequently visited but high-influence states and the trajectory leading to them. To this end, we sample the target influence value according to the log-probability:
\begin{equation}
    \mathcal{I}_i = \log p(\mathcal{I}|s_i) + \kappa (\mathcal{I} - \mathcal{I}_{low}) / (\mathcal{I}_{high} - \mathcal{I}_{low}),
\end{equation}
where $\kappa$ is a hyperparameter that controls the preference on high-influence states, $\mathcal{I}_{low}$ and $\mathcal{I}_{high}$ are the lower bound and upper bound of $\mathcal{I}_i$, respectively. The target timestep-to-go and return-to-go are obtained in the same way.

In addition, we also provide a one-shot demonstration for the imagination to avoid it being too far away from the target trajectory, also known as the ``hallucination'' problem in the large language model~\citep{mckenna2023sources,manakul2023selfcheckgpt}. Specifically, we store the trajectory segments in a few-shot demonstration dataset and use the prompt to characterize the segments. Before imagination, we search for the trajectory whose description has the highest similarity with current prompt $\mathcal{P}(s_i)$ and then prepend it into the original input $x$. This process only affects the inference procedure of the model – training remains unaffected and can rely on standard next-token prediction frameworks and infrastructure.
 
\subsection{Initialize and Explore} 
Before exploration, we begin by sampling a prompt $\mathcal{P}(s_0)=\{\mathcal{I},\mathcal{T},\mathcal{R}\}$ for imagination given the initial state $s_0$ from the environment simulator. Then, the agents imagine how to reach the interaction state $s_{\mathcal{T}}$ in an autoregressive manner by conditioning on the prompt and the most related trajectory from the few-shot demonstration dataset. After rolling out each imagination step and obtaining the next state, we have the next prompt through decrementing the target timestep-to-go $\mathcal{T}$ and return-to-go $\mathcal{R}$ by $1$ and the predicted reward $r_t$ from the imagination model, respectively. We maintain the influence value $\mathcal{I}$ as a constant and repeat this process until the target timestep-to-go is zero. The imagined trajectory $x_{0:\mathcal{T}}=\{s_0,o_0,\textbf{u}_0,r_0,s_1,...,\textbf{u}_{\mathcal{T}},r_{\mathcal{T}}\}$ is then used to initialize GRU state of the agent network. 

Next, we define a probability $\alpha$ of initializing the agents at state $s_0$ to emphasize the importance of the target task. With the probability $1-\alpha$, we initialize the agents at the last state $s_\mathcal{T}$ of the imagined trajectory using the simulator. For StarCraft II, we add a distribution over team unit types, start positions, and health points in the reset function of the environment simulator. We anneal $\alpha$ from $1.0$ to $0.5$ over a fixed number of steps after the pretraining phase of the imagination model. Finally, the agents interact with the environment to collect the online data $x_{\mathcal{T}:T}=\{s_\mathcal{T},o_\mathcal{T},\textbf{u}_\mathcal{T},r_\mathcal{T},s_{\mathcal{T}+1},...,\textbf{u}_{T},r_{T}\}$. We stitch the imagined trajectory and the online data $\mathcal{X}=x_{0:\mathcal{T}-1}\bigoplus x_{\mathcal{T}:T}$ as training datasets for policy training. 

The individual $Q$-value functions $Q^a(o^a,\tau^a,u^a)$ are optimized jointly to minimize the following loss:
\begin{equation}
    \min_{\theta,\phi} \sum_{t=0}^{T-1} \left[Q_{j}(s_t,\boldsymbol{\tau}_t,\mathbf{u}_t;\theta,\phi)-y(s_t,\boldsymbol{\tau}_t,\mathbf{u}_t)\right]^2,
\end{equation}
where $y(s_t,\boldsymbol{\tau}_t,\mathbf{u}_t)=r_t+\gamma \max_{\mathbf{u}}Q_{j}'(s_{t+1},\boldsymbol{\tau}_{t+1},\mathbf{u})$ is the target value function, and $Q_{j}'$ is the target network whose parameters are periodically copied from $Q_{j}$.

\section{Related Work} 

\begin{figure*}[t]
    \centering
    \includegraphics[width=0.94\linewidth]{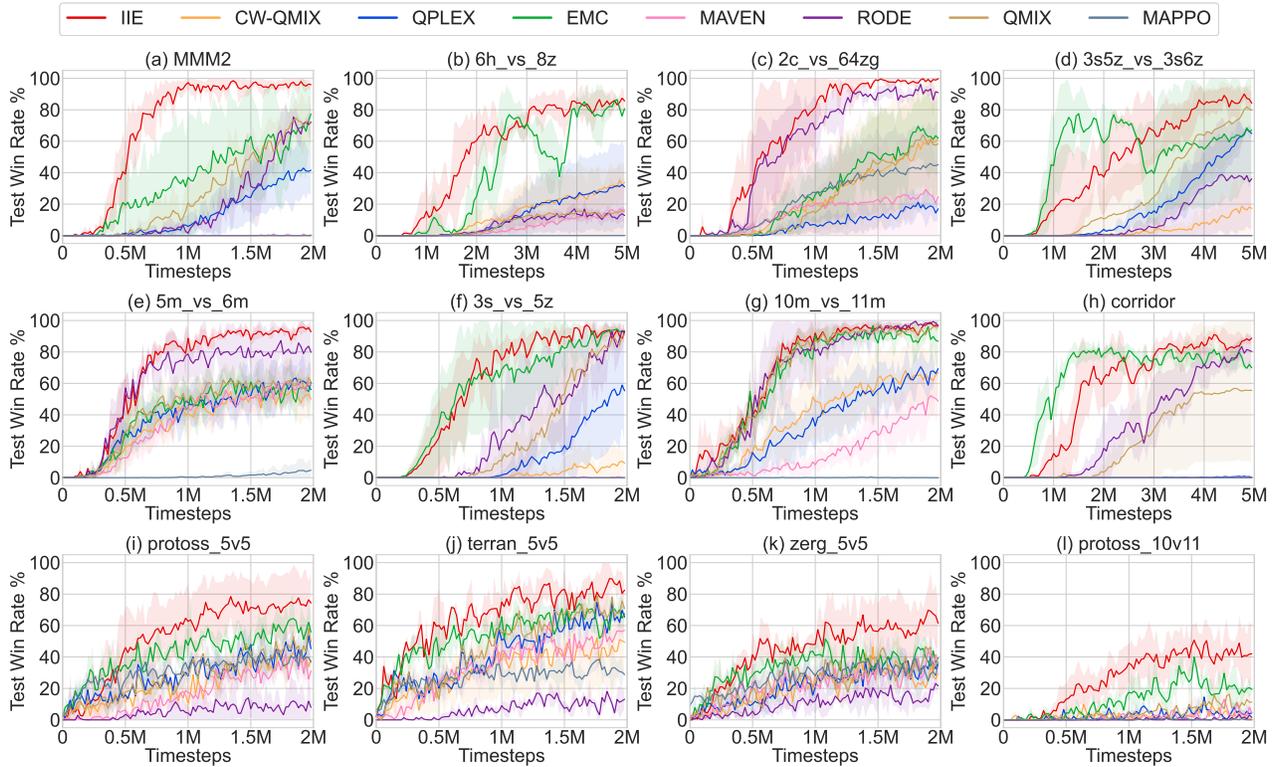}
    \caption{Performance comparisons on the dense-reward SMAC and SMACv2 benchmarks.}
    \label{smac}
\end{figure*}

\paragraph{Multi-agent Exploration.} Individual exploration suffers from the inconsistency between local and global information as well as the non-stationary problem in multi-agent settings. To address these limitations, \citet{jaques2019social} introduce intrinsic rewards based on ``social influence'' to incentivize agents in selecting actions that can influence other agents. Similarly, \citet{Wang2020Influence} leverage mutual information to capture the interdependencies of rewards and transitions, promoting exploration efficiency and facilitating the policies training. EMC~\citep{zheng2021episodic} leverages prediction errors from individual $Q$-values as intrinsic rewards and uses episodic memory to improve coordinated exploration. MAVEN~\citep{mahajan2019maven} employs a hierarchical policy for committed and temporally extended exploration, learning multiple state-action value functions for each agent through a shared latent variable. RODE~\citep{wang2020rode} introduces a role selector to enable informed decisions and learn role-based policies in a smaller action space based on the actions' effects. However, learning in long-horizon coordination tasks remains challenging due to the exponential state-action spaces.

\paragraph{Go-Explore.} Go-Explore~\citep{ecoffet2019go} is one of the most famous exploration approaches, particularly well-suited for hard-exploration domains with sparse or deceptive rewards. In the Go-Explore family of algorithms~\citep{ecoffet2021first,guo2020memory}, an agent returns to a promising state without exploration by running a goal-conditioned policy or restoring the simulator state and then explores from this state. However, these methods are primarily designed for single-agent scenarios, neglecting the interdependencies between agent policies and the strategies of others in MARL. Moreover, these methods require restoring previously visited simulator states or training a goal-conditioned policy to generate actions for returning. This leads to significant computational costs, especially in coordination scenarios involving complex observations and transitions.

\paragraph{Transformer Model.} Several works have explored the integration of transformer models into reinforcement learning (RL) settings. We classify them into two major categories depending on the usage pattern. The first category focuses on representing components in RL algorithms, such as policies and value functions~\citep{parisotto2020stabilizing,parisotto2021efficient}. These methods rely on standard RL algorithmis to update policy, where the transformer only provides the large representation capacity and improves feature extraction. Conversely, the second category aims to replace the RL pipeline with sequence modeling. They autoregressively generate states, actions, and rewards by conditioning on the desired return-to-go during inference~\citep{chen2021decision,lee2022multi,reed2022generalist}. However, there is a risk of unintended behaviors when offline datasets contain destructive biases. Moreover, it remains an open question of how to extend these supervised learning paradigms to online settings and satisfy the scalability in practice.

We utilize a transformer model to imagine the trajectory to reach the interaction state guided by a prompt, offering increased representational capacity and stability compared to existing Go-Explore methods. In contrast to current multi-agent exploration methods, we initialize agents at the states with high influence values. This form of curriculum learning significantly reduces the exploration space and enhances coordination exploration. We aim to bridge sequence modeling and transformers with MARL rather than replacing conventional RL algorithms.

\section{Results}
In this section, we conduct empirical experiments to answer the following questions: (1) Is Imagine, Initialize, and Explore (IIE) better than the existing MARL exploration methods in complex cooperative scenarios or sparse reward settings? (2) Can IIE generate a reasonable curriculum of the last states over timesteps and outperform other returning methods? We also investigate the contribution of each component in the proposed prompt to the imagination model.

We conduct experiments on NVIDIA RTX 3090 GPUs. Each task needs to train for about 12 to 20 hours, depending on the number of agents and the episode length limit. We evaluate 32 episodes with decentralized greedy action selection every $10k$ timesteps for each algorithm. All figures are plotted using mean and standard deviation with confidence internal 95\%. We conduct five independent runs with different random seeds for each learning curve.

\begin{figure*}[t]
    \centering
    \includegraphics[width=0.94\linewidth]{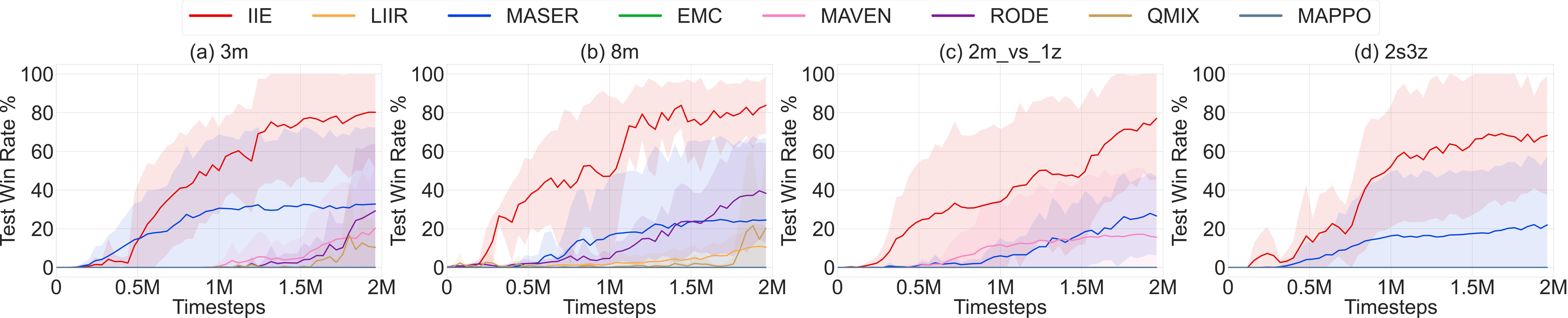}
    \caption{Performance comparisons on the sparse-reward SMAC benchmark.}
    \label{smac_sparse}
\end{figure*}
\begin{figure*}[t]
    \centering
    \includegraphics[width=0.94\linewidth]{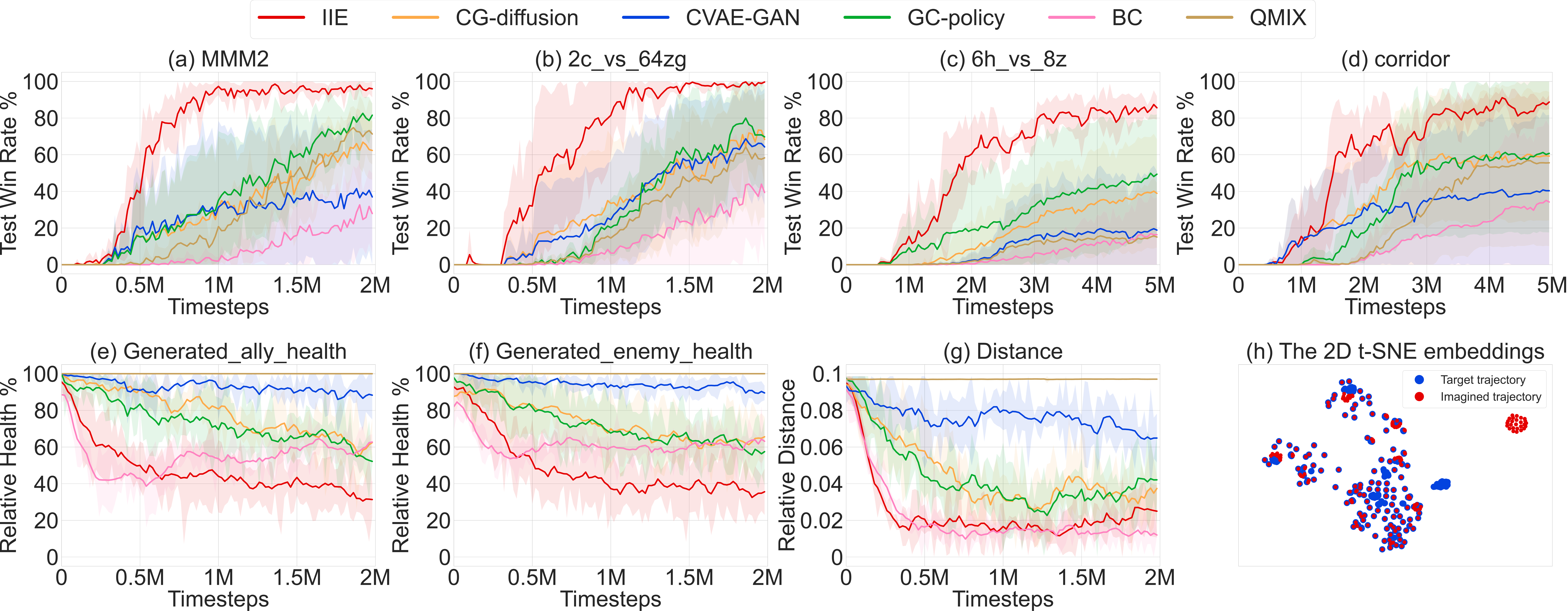} 
    \caption{(a-d) Performance comparisons with different returning methods on the SMAC benchmark. (e-g) The mean health of allies and enemies, as well as the relative distance between two groups at the last state in the \texttt{MMM2} scenario. (h) The 2D t-SNE embeddings of the trajectory returned from IIE in the \texttt{MMM2} scenario after pretraining.}
    \label{gene}
\end{figure*}
\begin{figure*}[t]
    \centering
    \includegraphics[width=0.94\linewidth]{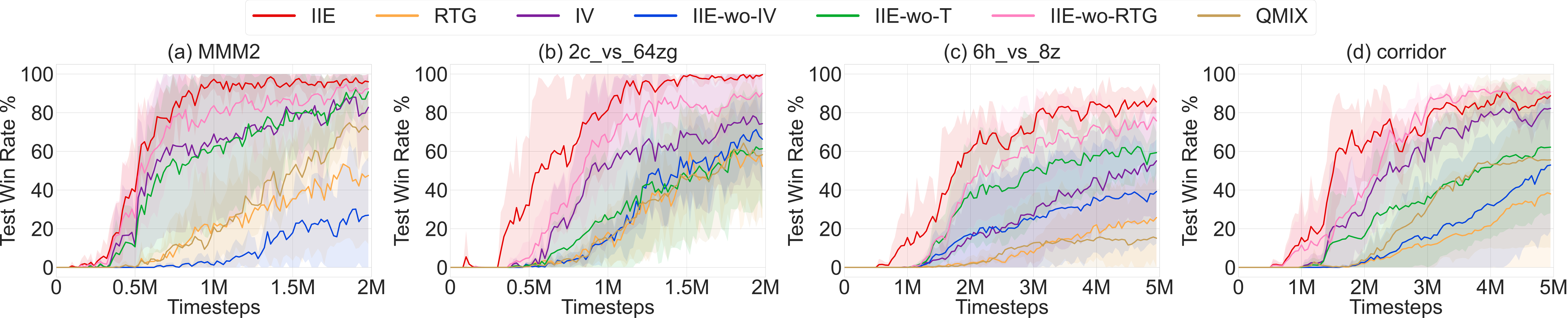}
    \caption{IIE with different prompts on the SMAC benchmark.}
    \label{prop}
\end{figure*}

\subsection{Performance Comparison}
In our evaluation, we compare the performance of CW-QMIX~\citep{rashid2020weighted}, QPLEX~\citep{wang2020qplex}, MAVEN~\citep{mahajan2019maven}, EMC~\citep{zheng2021episodic}, RODE~\citep{wang2020rode}, QMIX~\citep{rashid2018qmix}, MAPPO~\citep{yu2021surprising}, and IIE on the StarCraft Multi-Agent Challenge (SMAC)~\citep{samvelyan2019starcraft} and SMACv2 ~\citep{ellis2022smacv2} benchmarks. 

SMAC is a partially observable MARL benchmark known for its rich environments and high control complexity. It requires learning policies in a large observation-action space, where agents take various actions, such as ``move'' in cardinal directions, ``stop'', and selecting an enemy to attack. The maximum number of actions and the episode length vary across different scenarios, ranging from 7 to 70 actions and 60 to 400 timesteps, respectively. In contrast, SMACv2 presents additional challenges of stochasticity and generalization. It includes procedurally generated scenarios with start positions, unit types, attack range, and sight range. The agents must generalize their learned policies to previously unseen settings during testing.

Fig.~\ref{smac} shows that IIE considerably outperforms the state-of-the-art MARL methods in both SMAC and SMACv2 maps with the dense reward setting. This result demonstrates that IIE significantly enhances learning speed and coordination performance in complex tasks. In specific scenarios like \texttt{3s5z\_vs\_3s6z} and \texttt{corridor}, EMC shows faster learning in the beginning, which can be attributed to the fact that IIE requires more time to pre-train the imagination model in more complex tasks. However, as the training progresses, IIE excels in providing a more targeted and efficient exploration of complex coordination scenarios, leading to the best final performance across all scenarios.

\subsection{Sparse-reward Benchmark}
We investigate the performance of IIE, EMC, MAVEN, RODE, QMIX, and MAPPO, in addition to two multi-agent exploration methods designed for sparse rewards, including LIIR~\citep{du2019liir} and MASER~\citep{maserJeon2022}, on the SMAC benchmark with the sparse-reward setting. In this setting, global rewards are sparsely given only when one or all enemies are defeated, with no additional reward for state information such as enemy and ally health. This problem is difficult for credit assignments because credit must be propagated from the end of the episode.

The imagination model in IIE can improve robustness in these settings because it makes minimal assumptions on the density of the reward. As shown in Fig.~\ref{smac_sparse}, sparse and delayed rewards minimally affect IIE. We hypothesize that the transformer architecture in the imagination model can be effective critics and enable more accurate value prediction. In contrast, QMIX and MAPPO fail to solve these tasks since they heavily rely on densely populated rewards for calculating temporal difference targets or generalized advantage estimates. LIIR, MASER, RODE, and MAVEN exhibit slow learning and instability, particularly on the heterogeneous map \texttt{2s3z}, suggesting the difficulty of learning such intrinsic rewards, subgoals, roles, or noise-based hierarchical policies in hard-exploration scenarios. 

\subsection{Different Returning Methods}\label{visual}
In this section, we seek insight into whether the imagination model in IIE can be thought of as performing efficient curriculum learning and is better than other returning methods. To investigate this, we compare it with behavior cloning (BC), a goal-conditioned policy method (GC-policy)~\citep{ecoffet2021first}, and generative models, including CVAE-GAN~\cite{bao2017cvae} and classifier-guided diffusion (CG-diffusion)~\citep{ajay2023is}. Before exploration, we initialize the agents to the last state of the imagination trajectory produced by BC, CVAE-GAN, and CG-diffusion using the environment simulator or run a goal-conditioned policy from GC-policy to return agents to the target state.

We show the performance comparison and visualization results in Fig.~\ref{gene}. IIE provides agents with interaction points for exploration with the shortest relative distance between agents and enemies and decreasing but comparable health levels as training progresses. This contributes to early and more frequent multi-agent interactions against the enemies quickly, reducing the complexity of the multi-agent exploration space. As a result, IIE outperforms BC, GC-policy, CVAE-GAN, and CG-diffusion across all tasks, indicating that the prompt-based imagination can be more effective and has better generalization than simply performing imitation learning or other generative models. We also illustrate the data distribution of imagined trajectories in a 2D space with dimensional reduction via T-SNE. The results show that the imagined trajectory will be close to the target trajectory, which verifies that IIE can capture and learn the dynamics of the multi-agent environment based on the prompt.

CG-diffusion and GC-policy have plateaued in performance, showing a similar trend of emergent complexity with IIE but taking far longer to learn the joint policy. On the one hand, the continuous diffusion models have been extremely successful in vision and audio domains, but they do not perform well in text because of the inherently discrete nature of text~\citep{li2022diffusion}. The imagination of the trajectory is more related to text than image generation because it has low redundancy, and the transition between two connected states is essential. On the other hand, the applicability of the current goal-conditioned methods with complex observations is limited, as their flexibility is often constrained to task spaces using low-dimensional parameters and requires well-defined task similarity. CVAE-GAN does not show positive results and keeps generating similar health levels. We hypothesize that the mode collapse problem is a bottleneck for CVAE-GAN. The generator can find data that can easily fool the discriminator because of the unbalanced training data from the exploration. BC shows the worst results because it imitates all past interaction sequences and lacks the generalization ability to avoid sub-optimal solutions. From Fig.~\ref{gene}e-g, we can see that BC prioritizes states with short distances but does not provide reasonable health levels for enemies - the mean health level of enemies far exceeds that of allies, making it difficult or even impossible to achieve any success. 

\subsection{Different Prompts for Imagination}
In this section, we conduct ablation studies to analyze the contributions of each component in the prompts, including the timesteps, the accumulated return-to-go (RTG), the constant influence value (IV), and the most related trajectory (T). We integrate the desired timesteps into the following prompts: (1) RTG, (2) IV, (3) RTG with the influence value, denoted as IIE-wo-T, (4) RTG with the most related trajectory, denoted as IIE-wo-IV, (5) IV with the most related trajectory, denoted as IIE-wo-RTG. We compare them with IIE and QMIX on the SMAC benchmark.

Fig.~\ref{prop} shows that RTG and IIE-wo-IV achieve poor performance in \texttt{MMM2}, \texttt{6h\_vs\_8z}, and \texttt{corridor} due to their ignorance of the importance of interactions in multi-agent exploration and limited positive samples for value decomposition. IV and IIE-wo-T perform worse than IIE-wo-IV and IIE with a considerable gap, as they do not exploit one-shot demonstration to guide action generation, leading to a potential risk that the imagined trajectories may deviate significantly from the target trajectory, especially in long-horizon tasks with complex transition functions. IIE outperforms IIE-wo-IV across \texttt{MMM2}, \texttt{6h\_vs\_8z}, and \texttt{2c\_vs\_64zg} maps because RTG can further specify a trajectory, improving the efficiency and robustness of the imagination learning. However, IIE-wo-IV performs better than IIE in \texttt{corridor}, implying that the imagination model may have some degree of generalization without RTG and provide more diverse samples for policy training. 

\section{Conclusion}
We proposed Imagine, Initialize, and Explore, which enables us to break down a difficult multi-agent exploration problem into a curriculum of subtasks created by initializing agents at the interaction state - the last state in the imagined trajectory. We empirically evaluated our algorithm and found it outperforms current multi-agent exploration methods and generative models on various benchmarks. We also show that the prompt-based imagination model performs efficient conditional sequence generation and has one-shot generalization. We hope this work will inspire more investigation of sequence-prediction models' applications in multi-agent reinforcement learning (MARL) rather than using them to replace conventional MARL. In future work, we consider learning continuous prompts to specify the imagination, as opposed to the simple influence value used in this paper. 

\section{Acknowledgments}
This work was supported in part by National Key R\&D Program of China under grant No. 2021ZD0112700, NSFC under grant No. 62125305, No. 62088102, No. 61973246, No. 62203348, and No. U23A20339, the Fundamental Research Funds for the Central Universities under Grant xtr072022001.

\bigskip
\bibliography{aaai24}

\clearpage

\section{Appendix}

\subsection{Environments}
StarCraft II is a real-time strategy game featuring three different races, Protoss, Terran, and Zerg, with different properties and associated strategies. The objective is to build an army powerful enough to destroy the enemy’s base. When battling two armies, players must ensure army units are acting optimally.

StarCraft Multi-Agent Challenge (SMAC) is a partially observable reinforcement learning benchmark built in StarCraft II. An individual agent with parameter sharing controls each allied unit, and a hand-coded built-in StarCraft II AI controls enemy units. The difficulty of the game AI is set to the ``very difficult'' level. On the SMAC benchmark, agents can access their local observations within the field of view at each time step. The feature vector contains attributes of both allied and enemy units: \texttt{distance}, \texttt{relative\ x}, \texttt{relative\ y}, \texttt{health}, \texttt{shield}, and \texttt{unit\_type}. In addition, agents can observe the last actions of allied units and the terrain features surrounding them. The global state vector includes the coordinates of all agents relative to the center of the map and other features present in the local observation of agents. The state stores the energy of Medivacs, the cooldown of the rest of the allied units, and the last actions of all agents. Note that the global state information is only available to agents during centralized training. All features in state and local observations are normalized by their maximum values. 

After receiving the observations, each agent is allowed to take action from a discrete set which consists of \texttt{move[direction]}, \texttt{attack[enemy\_id]}, \texttt{stop} and \texttt{no-op}. Move direction includes north, south, east, and west. Note that the dead agents can only take \texttt{no-op} action while live agents cannot. For health units, Medivacs use \texttt{heal[agent\_id]} actions instead of \texttt{attack[enemy\_id]}. Depending on different scenarios, the maximum number of actions varies between $7$ and $70$. Note that agents can only perform the \texttt{attack[enemy\_id]} action when the enemy is within its shooting range. At each time step, agents take joint action and receive a positive global reward based on the total damage dealt to the enemy units. In addition, they can receive an extra reward of $10$ points after killing each enemy unit and $200$ points after killing all enemy units. The rewards are scaled to around $20$, so the maximum cumulative reward is achievable in each scenario.

\begin{table}[h]
	\centering
	\begin{tabular}{c|cc}
		\toprule 
		Attribution & Dense reward & Sparse reward  \\
		\midrule 
		Win & +200 & +200 \\
        One enemy dies & +10 & +10 \\
        One ally dies & 0 & 0\\
        Enemy's health & + health difference & 0 \\
        Enemy's shield & + shield difference & 0 \\
        Ally's health & 0 & 0 \\
        Enemy's shield & 0 & 0 \\
        \bottomrule
	\end{tabular}%
    \caption{Reward setting.}
	\label{tab_reward}%
\end{table}%

SMACv2 is a new version of SMAC that uses procedural content generation to improve stochasticity, including random team compositions, random start positions, and increasing diversity among unit types by using the true unit attack and sight ranges. Each race has a special unit that should not be generated too often: the colossus in Protoss, the medivac unit in Terran, and the baneling unit in Zerg. All of these special units are spawned with a probability of 10\%. The other units used spawn with a probability of 45\%. Random start positions in SMACv2 come in two different flavors. In reflect scenarios, the allied units are spawned uniformly randomly on one side of the scenario. The enemy positions are the allied positions reflected in the vertical midpoint of the scenario. This is similar to how units spawned in the original SMAC benchmark but without clustering them together. In surround scenarios, allied units are spawned at the center and surrounded by enemies stationed along the four diagonals. There are two changes to the observation space from SMAC. First, each agent observes their field-of-view direction. Secondly, each agent observes their position in the map as x- and y-coordinates. This is normalized by dividing by the map width and height, respectively. The only change to the state from SMAC was to add the field-of-view direction of each agent to the state. In addition, the fixed attack and sight range are replaced by the values from SC2. The attack range for melee units is imposed to 2 because using the actual attack ranges makes attacking too difficult.

The reward setting for the dense and sparse cases on the SMAC benchmark is shown in Tab.~\ref{tab_reward}.

We use \texttt{SC2.4.6.2.69232} (the same version for the evaluation of VDN, QMIX, QPLEX, QTRAN, and RODE) instead of \texttt{SC2.4.10} (the version for the evaluation of MAPPO) and \texttt{SC2.4.1.4} (the version for the evaluation of LIIR). Performance is \textbf{not} comparable across versions.

\subsection{Experimental Setup}\label{pro_dis}
We adopt the same architectures for QMIX$^1$, QPLEX$^1$, CW-QMIX\footnote{https://github.com/oxwhirl/wqmix}, RODE\footnote{https://github.com/TonghanWang/RODE}, MAVEN\footnote{https://github.com/AnujMahajanOxf/MAVEN}, EMC\footnote{https://github.com/kikojay/EMC} as their official implementations~\citep{samvelyan2019starcraft,wang2020qplex,rashid2020weighted,wang2020rode,mahajan2019maven,zheng2021episodic}. 

Each agent independently learns a policy with fully shared parameters between all policies. We used RMSProp with a learning rate of $5\times 10^{-4}$ and $\gamma=0.99$, buffer size 5000, mini-batch size 32 for all algorithms. The dimension of each agent's GRU hidden state is set to 64. For our experiments, we employ an $\epsilon$-greedy exploration scheme for the joint policy, where $\epsilon$ decreases from 1 to 0.05 over 1 million timesteps in \texttt{6h\_vs\_8z}, \texttt{3s5z\_vs\_3s6z} and \texttt{corridor}, and over 50 thousand timesteps in other maps. 

The pretraining phase starts at the beginning of the training and ends at $50k$ and $100k$ for $2M$-step and $5M$-step maps, respectively. During the pretraining, we set the probability $\alpha$ of initializing the agents at state $s_0$ as $1$. After pretraining, the probability $\alpha$ annealed from $1.0$ to $0.5$ over $50k$ and $100k$ for $2M$-step and $5M$-step maps, respectively.

We build our imagination model implementation based on Decision Transformer\footnote{https://github.com/kzl/decision-transformer}~\citep{chen2021decision}. The full list of hyperparameters can be found in Tab.~\ref{dthyper}. We use the maximum timesteps $E_{t}^{M}$ in the environment as the context length, which is shown in Tab.~\ref{maxt}. The imagination models were trained using the AdamW optimizer. 

\begin{table}[t]
	\centering
	\begin{tabular}{cc|cc}
		\toprule 
		Scenario & $E_{t}^{M}$ & Scenario & $E_{t}^{M}$ \\
		\midrule 
		\textit{MMM2} & 180 & \textit{6h\_vs\_8z} & 150 \\
        \textit{2c\_vs\_64zg} & 400 & \textit{3s5z\_vs\_3s6z} & 170 \\
        \textit{5m\_vs\_6m} & 70 & \textit{10m\_vs\_11m} & 150\\
        \textit{corridor} & 400 & \textit{3s\_vs\_5z} & 250\\
        \textit{protoss\_5v5} & 200 & \textit{terran\_5v5} & 200 \\
        \textit{protoss\_10v11}& 200 & \textit{zerg\_5v5} & 200 \\
        \textit{3m}& 60 & \textit{8m} & 120 \\
        \textit{2m\_vs\_1z}& 150 & \textit{2s3z} & 120 \\
        \bottomrule
	\end{tabular}%
    \caption{The maximal timesteps in each map.}
	\label{maxt}%
\end{table}%

\begin{table}[t]
	\centering
	\begin{tabular}{cc|cc}
		\toprule 
		Hyperparameter & Value & Hyperparameter & Value \\
		\midrule 
		number of layers & 6 & max timesteps & 400 \\
        attention heads & 8 & weight decay & 0.1 \\
        embedding dims & 64 & max RTG & 20\\
        grad norm clip & 1.0 & learning rate & $6\times 10^{-4}$\\
        Adam betas & (0.9,0.95) & training epochs & 5 \\

        \bottomrule
	\end{tabular}%
    \caption{Hyper-parameters in the transformer-based imagination model and behavior cloning.}
	\label{dthyper}%
\end{table}%

The prompt generator uses a feedforward network with two hidden layers of size $\{64,32\}$ to encode the state and output $\{E_{t}^{M},20,10\}$ dimensional vectors for each modality, i.e., the target timestep, return-to-go, and influence value, respectively. These vectors then turn into class probabilities using a softmax function.

For CW-QMIX, the weight for negative samples is set to $\alpha=0.5$ for all scenarios.

For CG-diffusion, the implementation is based on Diffuser\footnote[1]{https://github.com/jannerm/diffuser}~\citep{janner2022planning}. We use $N=100$ diffusion steps for all scenarios.

For CVAE-GAN, the generator $G(s'|z,c)$ first encodes the $64$-dimensional input noise $z$ using a $64$-dimensional fully-connected layer, where $c=(s,\mathcal{P}(s))$ is the condition, Then, it produces the reconstructed next state by another fully-connected layer. We apply a sigmoid function at the output layer and scale the output by the range of each modality defined by the task space. Batch normalization is used in all layers. The classifier $C(\mathcal{P}(s)|s)$ uses the same architecture as the prompt generator in IIE and is trained in the same way. In the discriminator $D(G(s'|z,c))$, a feedforward network with two hidden layers of size $\{128,64\}$ is used to encode the state or the output from the generator, and then predict a score. The loss function for CVAE-GAN is the same as the objective in~\citep{bao2017cvae}.

For GC-policy, we treat any trajectory as a successful trail for reaching its final state conditioned on the prompt. Therefore, we train the individual policy for each agent $i$ by maximizing the likelihood of the actions for a reach goal $J(\pi_i)=\mathbb{E}_D [\log_{\pi_i} (a|s,\mathcal{P}(s))]$. This policy is parametrized using a neural network that takes as the input state and the goal (prompt), then returns probabilities for a discretized grid of actions of the action space. The neural network concatenates the state and goal together. It passes the concatenated input into a feedforward network with two hidden layers of size $256$ and $128$, respectively, outputting logits for each discretized action. The loss is optimized using the Adam optimizer with learning rate $\alpha=5\times10^{-4}$, with a batch size of 256 transitions, taking one gradient step for every step in the environment. In GC-policy, the agents reach the selected state from an archive using the goal-conditioned policy rather than the environment simulator. The archive is a first-in-first-out buffer that stores states with the top-$128$ influence value.

The implementation of MAPPO is consistent with their official repositories\footnote[2]{https://github.com/zoeyuchao/mappo}~\citep{yu2021surprising}. As shown in Tab.~\ref{tabl}, all hyper-parameters are left unchanged at the origin best-performing status.

For the baselines in the sparse-reward setting, we adopt the same architecture for LIIR\footnote[3]{https://github.com/yalidu/liir}~\citep{du2019liir} and MASER\footnote[4]{https://github.com/Jiwonjeon9603/MASER}~\citep{maserJeon2022} as their official implementations.

\begin{table}[t]
	\centering
	\begin{tabular}{cc|cc}
		\toprule 
		Hyperparameter & Value & Hyperparameter & Value \\
		\midrule 
		critic lr & 5e-4 & actor lr & 5e-4 \\
        ppo epoch & 5 & ppo-clip & 0.2 \\
        optimizer & Adam & batch size & 3200\\
        optim eps & 1e-5 & hidden layer & 1 \\
        gain & 0.01 & training threads & 32 \\
        rollout threads & 8 & $\gamma$ & 0.99 \\
        hidden layer dim & 64 & activation & ReLU \\
        \bottomrule
	\end{tabular}%
    \caption{Hyper-parameters in MAPPO.}
	\label{tabl}%
\end{table}%

We conduct experiments on an NVIDIA RTX 3090 GPU. Each task needs to train for about 12 to 20 hours, depending on the number of agents and episode length limit of each map. We evaluate 32 episodes with decentralized greedy action selection every $10k$ timesteps for each algorithm.

All figures in the experiments are plotted using mean and standard deviation with confidence internal 95\%. We conduct five independent runs with different random seeds for each learning curve.

\subsection{Pseudocode for Prompt Sampling}

\begin{algorithm}[h]
    \caption{Pseudocode for Prompt Sampling}
    \label{alg:sample}
    Given an environment $E$, the state $s_0$. $\kappa=10$. The return upper bound $\mathcal{R}_{high}=20$, the return lower bound $\mathcal{R}_{low}=0$, the timestep upper bound $\mathcal{T}_{max}=E_{t}^{M}$, the timestep lower bound $\mathcal{T}_{min}=0$, the influence upper bound $\mathcal{I}_{max}=10$, the influence lower bound $\mathcal{I}_{min}=0$\\
    \begin{algorithmic}[1] 
    \STATE Compute $11$ logits ($\mathcal{I}= 0,...,10$) for the categorical influence value distribution $p(\mathcal{I}|s_0)$ \\
    \textit{\# Increase logits proportionally to influence value magnitudes to prefer high-influence state in the trajectory}
    \STATE Define a log-probability $\log P(\mathcal{I}^*|s_0) = \log p(\mathcal{I}|s_0) + \kappa (\mathcal{I} - \mathcal{I}_{low}) / (\mathcal{I}_{high} - \mathcal{I}_{low})$\\
    \textit{\# Sample a influence value}
    \STATE $\mathcal{I}_0\sim P(\mathcal{T}^*|s_0) $
    \STATE Compute $E_{t}^{M}+1$ logits ($\mathcal{T}= 0,...,E_{t}^{M}$) for the categorical timestep distribution $p(\mathcal{T}|s_0,\mathcal{I}_0)$ \\
    \textit{\# Increase logits proportionally to timestep magnitudes to prefer the timestep later in the trajectory}
    \STATE Define a log-probability $\log P(\mathcal{T}^*|s_0,\mathcal{I}_0) = \log p(\mathcal{T}|s_0,\mathcal{I}_0) + \kappa (\mathcal{T} - \mathcal{T}_{low}) / (\mathcal{T}_{high} - \mathcal{T}_{low})$\\
    \textit{\# Sample a timestep}
    \STATE $\mathcal{T}_0\sim P(\mathcal{T}^*|s_0,\mathcal{I}_0) $
    \STATE Compute $21$ logits ($\mathcal{R}= 0,...,20$) for the categorical return distribution $p(\mathcal{R}|s_0,\mathcal{T}_0)$ \\
    \textit{\# Increase logits proportionally to return magnitudes to prefer high-return state in the trajectory}
    \STATE Define a log-probability $\log P(\mathcal{R}^*|s_0,\mathcal{I}_0,\mathcal{T}_0) = \log p(\mathcal{R}|s_0,\mathcal{I}_0,\mathcal{T}_0) + \kappa (\mathcal{R} - \mathcal{R}_{low}) / (\mathcal{R}_{high} - \mathcal{R}_{low})$\\
    \textit{\# Sample a return}
    \STATE $\mathcal{R}_0\sim P(\mathcal{R}^*|s_0,\mathcal{I}_0,\mathcal{T}_0) $
    \STATE \textbf{return} $\mathcal{I}_0,\mathcal{T}_0,\mathcal{R}_0$
    \end{algorithmic}
    \end{algorithm}

\subsection{Environment Simulator}\label{eses}

We design a convenient interface in the reset function, serving as the environment simulator in Imagine, Initialize, and Explore. Suppose all units are initialized at the starting point $s_0$, and we have already obtained the imagined trajectory $x_{0:\mathcal{T}}=\{s_0,o_0,\textbf{u}_0,r_0,s_1,...,\textbf{u}_{\mathcal{T}},r_{\mathcal{T}}\}$. The last state $s_{\mathcal{T}}$ is formulated as a distribution class $\mathcal{D}$ over team unit types, start positions, and health levels.

First, we kill all units through the \textit{DebugKillUnit} command from the \textit{s2clientprotocol.debug\_pb2} package. Then, we create new units based on the distributions $\mathcal{D}$ 
through the \textit{DebugCreateUnit} command, where we can set the \textit{unit\_type}, the \textit{owner} (allies or enemies), and initialized positions using the \textit{Point2D} command. Note that we have to recover the coordinates of all units in the real axis because they have been processed as the relative value to the center of the map in the state and the observations.

Since we cannot directly initialize the health level of a unit through the \textit{DebugCreateUnit} command, we scale the health value in the raw observation from the controller and kill agents when their health is below the health level. For example, consider the maximal health of the agent $i$ is $H_i$, the remaining health value at the last state $s_{\mathcal{T}}$ is $\beta H_i$. Suppose the exact health value in the raw observation is $h_i$. We compute the current health by $h_i^c=\left[h_i-(1-\beta)H_i\right]/H_i$ and use it as the feature of health values for the state and the observations. Moreover, we kill the agent $i$ if $h_i^c\leq 0$ after taking the joint action at each timestep.

    \subsection{Imagine, Initialize, Explore Pseudocode}
    \begin{algorithm}[h]
        \caption{Imagine, Initialize, Explore Pseudocode}
        \label{alg:algorithm}
        \textbf{Input}: initial agent network parameters $\theta$, initial mixing network parameters $\phi$, initial imagination model $\psi$, initial prompt generator $\xi$, replay buffer $B$, few-shot demonstrations $D$\\
        \textbf{Hyperparameters}: sampling ratio $\alpha$, segments number $K$\\
        \begin{algorithmic}[1] 
        \WHILE{not converged do}
        \STATE Reset the environment to the initial state $s_0$
        \STATE Sample $z\sim U(0,1)$ 
        \IF {$z > \alpha$}
        \STATE Sample a prompt $\mathcal{P}(s_i;\xi)=\{\mathcal{I}_0,\mathcal{T}_0,\mathcal{R}_0\}$ according to Algorithm~\ref{alg:sample}
        \STATE Prepend the one-shot demonstration whose discription has the highest similarity with the sampled prompt before the input of the imagination model\\
        \textit{\# Imagine}
        \STATE Generate the imagined trajectory through the imagined model\\
        \textit{\# Initialize}
        \STATE Reset the environment to the last state of the imagined trajectory
        \STATE Initialize GRU state of the agent network\\
        \textit{\# Explore}
        \STATE Explore until the environment is terminated, store the stitched trajectory into $B$
        
        \ELSE 
        \STATE Explore until the environment is terminated, store the interaction sequence into $B$
        \STATE Obtain $K$ trajectory segments and store them into $D$
        \STATE Update the prompt generator $\xi$ using segments
        \STATE Update the imagination model $\psi$ using segments
    
        \ENDIF
        \STATE Sample a batch of sequences from $B$
        \STATE Update the agent network $\theta$ and the mixing network $\phi$

        \ENDWHILE
        \end{algorithmic}
        \end{algorithm}

\subsection{More Related Work}
\paragraph{Auto-curricula Learning.} Curriculum learning is a technique that leverages easier datasets or tasks to facilitate training. In reinforcement learning (RL), this approach involves selecting tasks from a predefined set or parameterized space of goals and scenes to expedite performance improvement on the target task~\citep{florensa2018automatic,matheron2020pbcs,fang2021adaptive,ao2022eat}. In the multi-agent field, recent works have explored the adaptive selection of co-players in competitive games, where playing against increasingly stronger opponents is crucial to avoid exploitation by other agents~\citep{silver2018general,samvelyan2023maestro}. These methods use a regret-based curriculum to attain a Nash-Equilibrium policy against every rational agent in every environment. It is important to note that these approaches primarily concentrate on competitive tasks rather than cooperative ones. Additionally, the applicability of the current auto-curricula learning methods to MARL tasks with complex observations is limited, as their flexibility is often constrained to task spaces using low-dimensional parameters and requires well-defined task similarity.

\subsection{Limitations}
The main limitation of this paper is that it is confined to scenarios within the game StarCraft II. This is an environment that, while complex, cannot represent the dynamics of all multi-agent tasks. Evaluation of MARL algorithms, therefore, should not be limited to one benchmark but should target a variety with a range of tasks.

The environment simulator in this paper is a small distribution class that can change how units are generated. It can be easily generalized to many multi-agent benchmarks and applications. For example, we can add a configuration of the initial position of the agents into the \textit{reset\_world} function on the particle world benchmark~\citep{lowe2017multi}. We can also utilize the \textit{AddPlayer} function to build a player by defining its coordinates and role on the Google Research Football benchmark~\citep{kurach2020google}. We leave the implementation of this functionality as future work.

\end{document}